\pdfoutput=1

\documentclass[11pt]{article}

\usepackage[]{acl}

\usepackage{times}
\usepackage{latexsym}

\usepackage[T1]{fontenc}

\usepackage[utf8]{inputenc}

\usepackage{microtype}
\usepackage{algorithm}
\usepackage{algorithmic}
\usepackage{multirow}
\usepackage{comment}
\usepackage{amsmath}
\usepackage{xcolor}
\usepackage{amsthm}
\usepackage{booktabs}
\usepackage{graphicx}

\title{Including Facial Expressions in Contextual Embeddings for Sign Language Generation}

\author{
    Carla Viegas\textsuperscript{\rm 1,}\textsuperscript{\rm 2} \and
    Mert Inan\textsuperscript{\rm 3} \and
    Lorna Quandt\textsuperscript{\rm 4} \and
    Malihe Alikhani \textsuperscript{\rm 3} \\
    \textsuperscript{\rm 1}Stella AI, Pittsburgh, USA\\
    \textsuperscript{\rm 2}Language Technologies Institute, Carnegie Mellon University, Pittsburgh, USA\\
    \textsuperscript{\rm 3} Computer Science Department, School of Computing and Information, \\
    University of Pittsburgh, Pittsburgh, USA\\
    \textsuperscript{\rm 4} Educational Neuroscience Program, Gallaudet University, Washington, D.C, USA \\
}
    
\begin{document}
\newtheoremstyle{break}
  {\topsep}{\topsep}%
  {\itshape}{}%
  {\bfseries}{}%
  {\newline}{}%
\theoremstyle{break}
\newtheorem{exmp}{Example}

\maketitle

\begin{abstract}
State-of-the-art sign language generation frameworks lack expressivity and naturalness which is the result of only focusing manual signs, neglecting the affective, grammatical and semantic functions of facial expressions. The purpose of this work is to 
augment semantic representation of sign language through grounding facial expressions. 
We study the effect of modeling the relationship between text, gloss, and facial expressions on the performance of the sign generation systems. In particular, we propose a Dual Encoder Transformer able to generate manual signs as well as facial expressions by capturing the similarities and differences found in text and sign gloss annotation. We take into consideration the role of facial muscle activity to express intensities of manual signs by being the first to employ facial action units in sign language generation. We perform a series of experiments showing that our proposed model improves the quality of automatically generated sign language.
\end{abstract}

\section{Introduction}


Communication between the Deaf and Hard of Hearing (DHH) people and hearing non-signing people may be facilitated by the emerging language technologies.
DHH individuals are medically underserved  worldwide~\cite{mckee2020overcoming, masuku2021world} due to the lack of doctors who can understand and use sign language. Also, educational resources that are available in sign language are limited especially in STEM fields~\cite{pittir40394,lynn2020successes}. Although the Americans with Disabilities Act~\cite{ADA} requires government services, public accommodations, and commercial facilities to communicate effectively with DHH individuals, the reality is far from ideal. Sign language interpreters are not always available and communicating through text is not always feasible as written languages are completely different from signed languages. 

\begin{figure*}[!ht]
    \centering
    \includegraphics[scale=0.38]{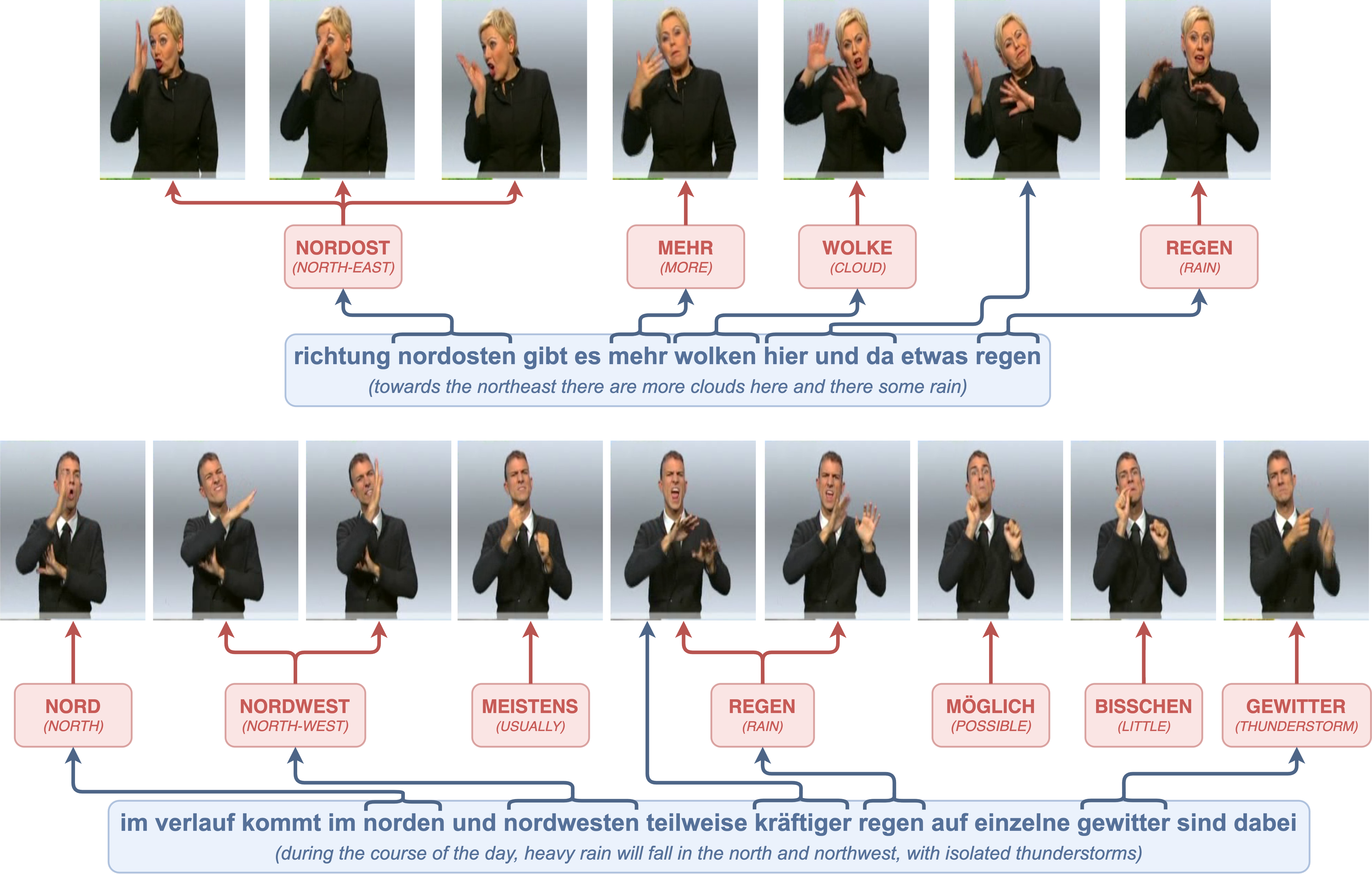}
    \caption{Sign Language uses multiple modalities, such as hands, body, and facial expressions to convey semantic information. Although gloss annotation is often used to transcribe sign language, the above examples show that meaning encoded through facial expressions are not captured. In addition, the translation from text (blue) to gloss (red) is lossy even though sign languages have the capability to express the complete meaning from text. The lower example shows lowered brows and a wrinkled nose to add the meaning of \texttt{kräftiger\textit{(heavy)}} (present in text) to the \textsc{rain} sign.}
    \label{fig:main_fig}
\end{figure*}

In contrast to Sign Language Recognition (SLR) which has been studied for several decades~\cite{rastgoo2021sign} in the computer vision community~\cite{yin2021including}, Sign Language Generation (SLG) is a more recent and less explored research topic ~\cite{quandt2021attitudes,cox2002tessa,glauert2006vanessa}. 

Missing a rich grounded semantic representation, the existing SLG frameworks are far from generating understandable and natural sign language. Sign languages use spatiotemporal modalities and encode semantic information in manual signs and also in facial expressions. A major focus in SLG has been put on manual signs, neglecting the affective, grammatical, and semantic roles of facial expressions. In this work, 
we bring insights from computational linguistics
to study the role of facial expressions in automated SLG. Apart from using facial landmarks encoding the contours of the face, eyes, nose, and mouth, we are the first to explore the use of facial Action Units (AUs) to learn semantic spaces or representations for sign language generation. 

In addition, with insights from multimodal Transformer architecture design, we present a novel model, the Dual Encoder Transformer for SLG, which takes as input spoken text and glosses, computes the correlation between both inputs, and generates skeleton poses with facial landmarks and facial AUs. Previous work used either gloss or text to generate sign language or used text-to-gloss (T2G) prediction as an intermediary step~\cite{saunders2020progressive}. Our model architecture, on the other hand, allows us to capture information otherwise lost when using gloss only, and captures differences between text and gloss, which is especially useful for highlighting adjectives otherwise lost in gloss annotation. We perform several experiments using the PHOENIX14-T weather forecast dataset and show that our model performs better than baseline models using only gloss or text.

In summary, our main contributions are the following:
\begin{itemize}
    \item Novel Dual Encoder Transformer for SLG which captures information from text and gloss, as well as their relationship to generate continuous 3D sign pose sequences, facial landmarks, and facial action units.
    \item Use of facial action units to ground semantic representation in sign language.
\end{itemize}

\section{Background and Related Work}

More than 70 million Deaf and Hard of Hearing worldwide use one of 300 existing sign languages as their primary language~\cite{kozik_2020}. In this section, we explain the linguistic characteristics of sign languages, the importance of facial expressions to convey meaning, and elaborate on prior work in SLG.

\subsection{Sign Language Linguistics}
Sign languages are spatiotemporal languages and are articulated by using the hands, face, and other parts of the body, which need to be visible. In contrast to spoken languages which are oral-aural languages, sign languages are articulated in front of the top half of the body and around the head. No universal method such as the International Phonetic Alphabet (IPA) exists to capture the complexity of signs. Gloss annotation is often used to represent the meaning of signs in written form. Glosses do not provide any information about the execution of the sign, only about its meaning. Even more, as glosses use written languages rather than the sign language, they are a mere approximation of the sign's meaning, representing only one possible transcription. For that reason, glosses do not always represent the full meaning of signs as shown in Figure~\ref{fig:main_fig}.

Every sign can be broken into four manual characteristics: shape, location, movement, and orientation. Non-manual components such as mouth movements (mouthing), facial expressions, and body movements are other aspects of sign language phonology. In contrast to spoken languages, signing occurs simultaneously while vowels and consonants occur sequentially. Although the vocabulary size of ASL in dictionaries is around 15,000~\cite{spreadthesign} compared to approximately 170,000 in spoken English, the simultaneity of phonological components allows for a wide range of signs to describe slight differences of the same gloss. 

While in English various words describe largeness (big, large, huge, humongous, etc.) in ASL, there is one main sign for “large”: \textsc{big}. However, through modifications of facial expressions, mouthing, and the size of the sign, different levels of largeness can be expressed just as in a spoken language~\cite{asl_vocab}.
To communicate spoken concepts without a corresponding sign fingerspelling---a manual alphabet---is sometimes used.~\cite{baker2016linguistics}

\begin{table}[]
\centering
\begin{tabular}{lcccc}
\cmidrule[\heavyrulewidth]{2-5}
      & \textbf{NOUN} & \textbf{VERB} & \textbf{ADV} & \textbf{ADJ} \\ \midrule
gloss & 20927         & 6407          & 17718        & 648          \\ \midrule
TEXT  & 25952         & 7638          & 24755        & 5628   \\ \bottomrule
\end{tabular}
\caption{Occurrence of different Part-of-Speech (POS) in the sign gloss annotation and the German transcripts computed with Spacy~\cite{spacy2}. Although gloss annotations show fewer samples for all POS, the difference in the occurrence of adjectives is statistically significant with $p<0.05$.}
\label{tab:pos}
\end{table}

\subsection{Grammatical Facial Expressions}

Facial expressions are grammatical components of sign languages that encode semantic representations, which, when excluded leads to loss of meaning. Facial expressions in particular have an important role in distinguishing different types of sentences such as WH-questions, Yes/No questions, doubt, negations, affirmatives, conditional clauses, focus and relative clauses~\cite{da2020recognition}. The following example shows how the same gloss order can present a question or an affirmation~\cite{baker2016linguistics}: 

\begin{exmp}
Indopakistani Sign Language\\
a) \textsc{father car exist.}\\\vspace{0.2cm}
\textup{``(My) father has a car.''}\\ 
b) \textsc{father car exist?}\\
\textup{``Does (your/his) father have a car.''}\\
\end{exmp}

In this example, what makes sentence b) a question are raised eyebrows and a forward and/or downward movement of the head/chin in parallel to the manual signs.

\begin{figure}[h]
    \centering
    \includegraphics[width=7cm]{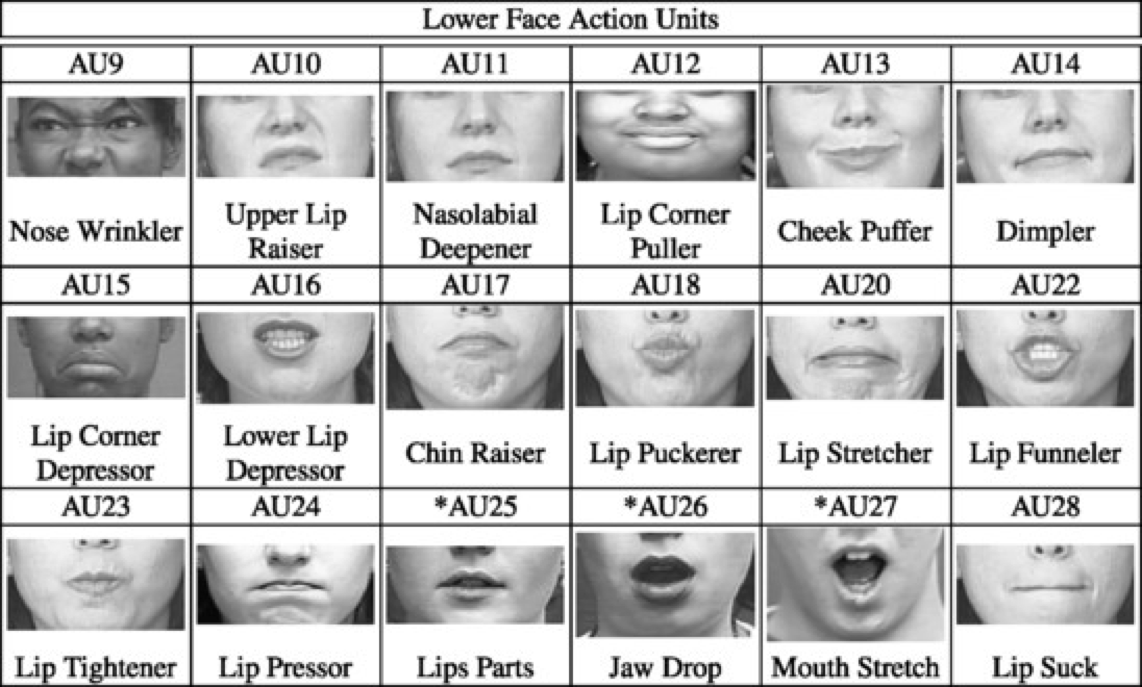}
    \caption{Examples from different facial Action Units (AUs)~\cite{friesen1978facial} from the lower face relevant to the generation of mouthings in sign languages. AUs can occur with different intensity values between 0 and 5. AUs have been used in psychology and in affective computing to understand emotions expressed through facial expressions. Image from~\cite{de2011facial}.}
    \label{fig:facs}
\end{figure}
In addition, facial expressions can differentiate the meaning of a sign assuming the role of a quantifier. Figure~\ref{fig:main_fig} shows different signs for the same gloss, \textsc{regen} (rain). We can observe from the text transcript (in blue) that the news anchor says ``rain'' in the upper example but ``heavy rain'' in the lower. This example shows how gloss annotations are not perfect transcriptions of sign languages as they only convey the meaning of manual aspect of the signs. Information conveyed through facial expressions to show intensities are not represented in gloss annotation. To view the loss of information that occurs in gloss annotation we used Spacy~\cite{spacy2} to compute the Part-of-Speech (POS) annotation for text and gloss. In Table~\ref{tab:pos} the occurrence of nouns, verbs, adverbs, and adjectives are shown for text and gloss over the entire dataset.
We can see that although gloss annotations have lower occurrence for all POS, the difference is statistically significant for adjectives with $p<0.05$. To calculate this significance, we performed hypothesis testing with two proportions by computing the Z score. We used t-tests to determine statistical significance of our model’s performance.

\begin{figure*}[ht]
    \centering
    \includegraphics[height=7.5cm]{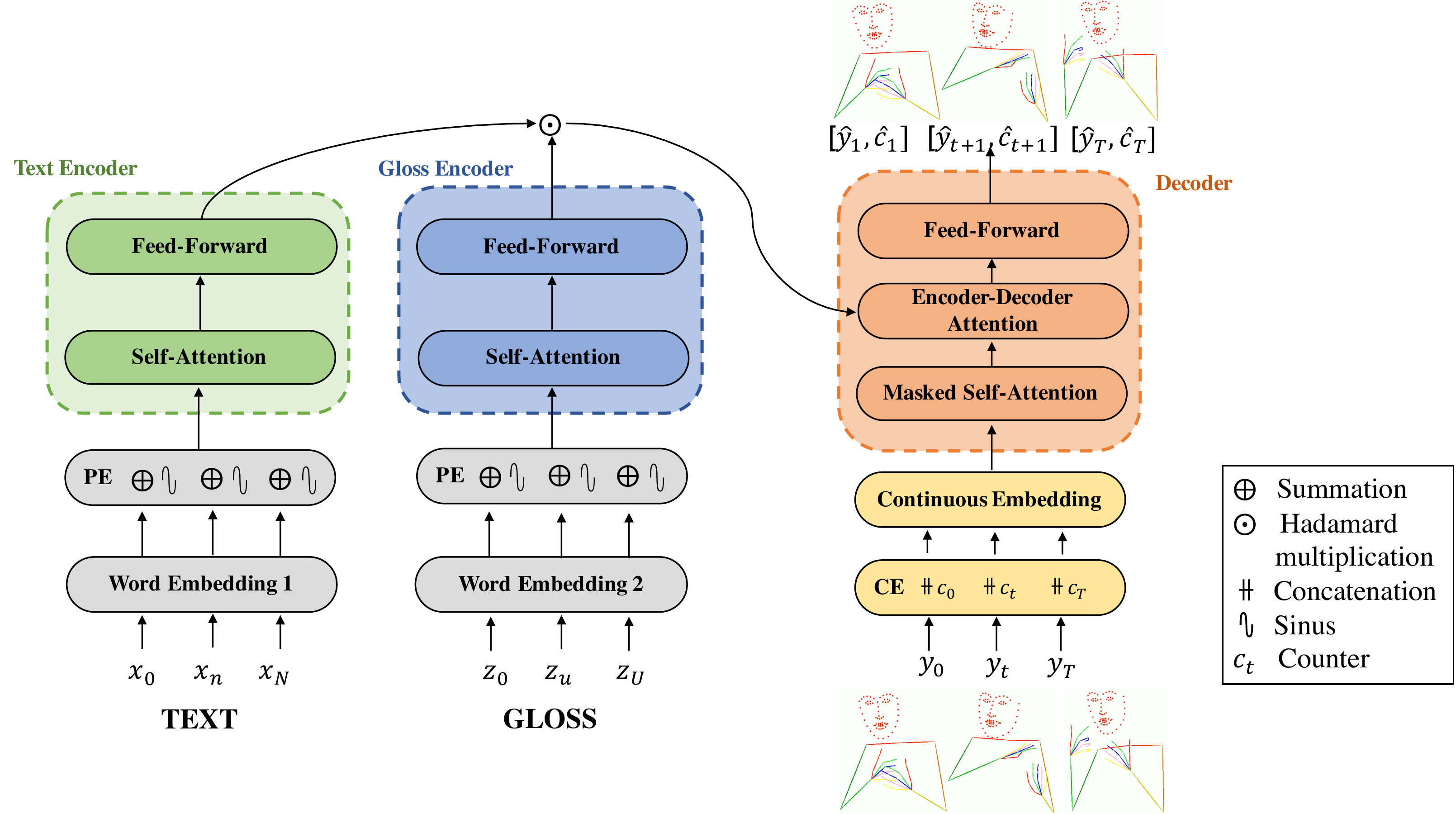}
    \caption{Our proposed model architecture, the Dual Encoder Transformer for Sign Language Generation. Our architecture is characterized by using two encoders, one for text and one for gloss annotation. The use of two encoders allows to multiply the outputs of both emphasizing the differences and similarities. In addition we to using skeleton poses and facial landmarks, we include facial action units~\cite{friesen1978facial}. }
    \label{fig:slt-tfn}
\end{figure*}

\subsection{Sign Language Generation}

Several advances in generating sign poses from text have been recently achieved in SLG, however there is limited work that considers the loss of semantic information when using gloss to generate poses and aligned facial expressions. Previous work has generated poses by translating text-to-gloss (T2G) and then gloss-to-pose (G2S) or by using either text or gloss as input~\cite{stoll2020text2sign, saunders2020progressive}. We propose a Dual Encoder Transformer for SLG which trains individual encoders for text and gloss, and combines the encoder's output to capture similarities and differences. 

In addition, the majority of previous work on SLG has focused mainly on manual signs~\cite{stoll2020text2sign, saunders2020progressive,zelinka2020neural,saunders2021mixed}.~\cite{saunders2021continuous} are the first to generate facial expressions and mouthing together with hand poses. The representation used for the non-manual channels is the same as for the hand gestures, namely coordinates of facial landmarks. In this work we explore the use of facial Action Units (AUs) (see Figure~\ref{fig:facs}) which represent intensities of facial muscle movements~\cite{friesen1978facial}. Although AUs have been primarily used in tasks related to emotion recognition~\cite{viegas2018towards}, recent works have shown that AUs help detect WH-questions, Y/N questions, and other types of sentences in Brazilian Sign Language~\cite{da2020recognition}.

\section{Sign Language Dataset}

In this work we use the publicly available PHOENIX14T dataset~\cite{camgoz2018neural}, frequently used as a benchmark dataset for SLR and SLG tasks. The dataset comprises a collection of weather forecast videos in German Sign Language (DGS), segmented into sentences and accompanied by German transcripts from the news anchor and sign-gloss annotations. PHOENIX14T contains videos of 9 different signers with 1066 different sign glosses and 2887 different German words. The video resolution is 210 by 260 pixels per frame and 30 frames per second. The dataset is partitioned into training, validation, and test set with respectively 7,096, 519, and 642 sentences.



\section{Methods: Dual Encoder Transformer for Sign Language Generation}

In this section, we present our proposed model, the Dual Encoder Transformer for Sign Language Generation.
Given the loss of information that occurs when translating from text-to-gloss, our novel architecture takes into account the information from text and gloss as well as their similarities and differences to generate sign language in the form of skeleton poses and facial landmarks shown in Figure~\ref{fig:slt-tfn}. 
For that purpose, we learn the conditional probability $p=(Y|X,Z)$ of producing a sequence of signs $Y=(y_1,\ldots,y_T)$ with $T$ frames, given the text of a spoken language sentence $X_T = (x_1,\ldots,x_N)$ with $N$ words and the corresponding glosses $Z=(z_1,\ldots,z_U)$ with $U$ glosses. 

Our work is inspired by the Progressive Transformer~\cite{saunders2020progressive} which allows translation from a symbolic representation (words or glosses) to a continuous domain (joint and face landmark coordinates), by employing positional encoding to permit the processing of inputs with varied lengths. In contrast to the Progressive Transformer which uses one encoder to use either text or glosses to generate skeleton poses, we employ two encoders, one for text and one for glosses, to capture information from both sources, and create a combined representation from the encoder outputs to represent correlations between text and glosses. In the following we will describe the different components of the dual encoder transformer.

\subsection{Embeddings}
As our input sources are words, we need to convert them into numerical representations. Similar to transformers used for text-to-text translations, we use word embeddings based on the vocabulary present in the training set. As we are using two encoders to represent similarities and differences between text and glosses we use one word embedding based on the vocabulary of the text and one using the vocabulary of the glosses. We also experiment by using the text word embedding for both encoders.
Given that our target is a sequence of skeleton joint coordinates, facial landmark coordinates, and continuous values of facial AUs with varying length we use counter encoding~\cite{saunders2020progressive}. The counter $c$ varies between [0,1] with intervals proportional to the sequence length. It allows the generation of frames without an end token. The target joints are then defined as:

\begin{align*}
    m_t &= [y_t,c_t]
    \textrm{ with } \\
    y_t &= [y_{hands+body}, y_{face}, y_{facialAUs}]
    \label{eq:embedding}
\end{align*}
    
The target joints $m_t$ are then passed to a continuous embedding which is a linear layer.

\subsection{Dual Encoders}
We use two encoders, one for text and one for gloss annotations. Both encoders have the same architecture. They are composed by $L$ layers each with one Multi-head Attention (MHA) and a feed-forward layer. Residual connections~\cite{he2016deep} around each of the two sublayers with subsequent layer normalization~\cite{ba2016layer}. MHA uses multiple projections of scaled dot-products which permits the model to associate each word of the input with each other.
The scaled dot-product attention outputs a vector of values, $V$, which is weighted by queries, $Q$, keys, $K$, and dimensionality, $d_k$:
\begin{equation}
    Attention(Q,K,V)=softmax(\frac{QK^T}{\sqrt{d_k}})
\end{equation} 

Different self-attention heads are used in MHA, which allows to generate parallel mappings of the $Q$, $V$, and $K$ with different learnt parameters.

The outputs of MHA are then fed into a non-linear feed-forward projection. In our case, where we employ two different encoders, their outputs can be formulated as:
\begin{equation}
\begin{split}
    H_n = E_{text}(\hat{w}_n, \hat{w}_{1:N})\\
    H_u = E_{gloss}(\hat{w}_u, \hat{w}_{1:U})
\end{split}    
\end{equation}
with $h_n$ being the contextual representation of the source sequence, $N$ the number of words, and $U$ the number of glosses in the source sequence.

As we want to not only use the information encoded in text and gloss, but also their relationship, we combine the output of both encoders with a Hadamard multiplication. As the $N\neq U$, we stack $h_n$ vertically for $U$ times and stack $h_u$ vertically for $N$ times in order to have two matrices with the same dimensions. Then we multiply both matrices with the Hadamard multiplication. Hadamard multiplication is a concatenation of every element in two matrices, where $a_{i,j}$ and $b_{i,j}$ are multiplied together to get $a_{i,j}b_{i,j}$. This represents concatenating the output vectors from the text encoder with the output of the vectors from the gloss encoder.

\begin{equation}
H_{text, gloss}=\begin{bmatrix}
H_{n0}\\
H_{n1} \\
\vdots\\
H_{nU}
\end{bmatrix} \odot
\begin{bmatrix}
H_{u0}\\
H_{u1} \\
\vdots\\
H_{uN}
\end{bmatrix}
\end{equation}

\subsection{Decoder}

Our decoder is based on the progressive transformer decoder (DPT), an auto-regressive model that produces continuous sequences of sign pose and the previously described counter value~\citep{saunders2020progressive}. In addition to producing sign poses and facial landmarks, our decoder also produces 17 facial AUs.
The counter-concatenated joint embeddings which include manual and facial features (facial landmarks and AUs), $\hat{j}_u$ , are used to represent the sign pose of each frame. Firstly, an initial MHA sub-layer is applied to the joint embeddings, similar to the encoder but with an extra masking operation. The masking of future frames is necessary to prevent the model from attending to future time steps.
A further MHA mechanism is then used to map the sym- bolic representations from the encoder to the continuous domain of the decoder. A final feed forward sub-layer follows, with each sub-layer followed by a residual connection and layer normalisation as in the encoder. The output of the progressive decoder can be formulated as:
\begin{equation}
    [\hat{y}_u, \hat{c}_u] = D(\hat{j}_{1:u-1}, h_{1:T})
\end{equation}

where $\hat{y}_u$ corresponds to the 3D joint positions, facial landmarks, and AUs, representing the produced sign pose of frame $u$ and $\hat{c}_u$ is the respective counter value. The decoder learns to generate one frame at a time until the predicted counter value, $\hat{c}_u$, reaches 1. The model is trained using the mean squared error (MSE) loss between the predicted sequence, $\hat{y}_{1:U}$ , and the ground truth, $y^{\ast}_{1:U}$ :
\begin{equation}
    L_{MSE} = \frac{1}{U}(y^{\ast}_{1:U}-\hat{y}_{1:U})^2
\end{equation}

\begin{table*}[!ht]
\small
\centering
\begin{tabular}{c|ccccc|ccccc}
\toprule
 \multirow{2}{*}{\textbf{Components}} & \multicolumn{5}{c|}{\textbf{Dev Set}} &  \multicolumn{5}{c}{\textbf{Test Set}} \\
& \multicolumn{1}{c}{Bleu$_1$} & \multicolumn{1}{c}{Bleu$_2$} & \multicolumn{1}{c}{Bleu$_3$} & \multicolumn{1}{c}{Bleu$_4$} & \multicolumn{1}{c|}{ROUGE} & \multicolumn{1}{c}{Bleu$_1$} & \multicolumn{1}{c}{Bleu$_2$} & \multicolumn{1}{c}{Bleu$_3$} & \multicolumn{1}{c}{Bleu$_4$} & \multicolumn{1}{c}{ROUGE} \\ 
\cmidrule[1pt]{2-11}
\begin{tabular}[c]{@{}c@{}}Manual\end{tabular} & 30.15 & 20.58 & 15.41 & 12.22 & 30.41 & 27.76 & 18.86 & 14.11 & 11.32 & 27.44 \\
\begin{tabular}[c]{@{}c@{}}Manual and Facial\end{tabular} & 29.46 & 20.30 & 15.31 & 12.10 & 29.25 & 26.75 & 17.88 & 13.29 & 10.61 & 26.54\\
\bottomrule
\end{tabular}
\caption{Translation results of the SLT model~\cite{camgoz2020sign} used for backtranslation when trained and evaluated with ground truth hand and body skeleton joints (manual) and facial landmarks and AUs (facial).}
\label{tab:slt}
\end{table*}

\begin{table*}[!ht]
\small
\centering
\begin{tabular}{c|ccccc|ccccc}
\toprule
 \multirow{2}{*}{\textbf{Model}} & \multicolumn{5}{c|}{\textbf{Dev Set}} &  \multicolumn{5}{c}{\textbf{Test Set}} \\
& \multicolumn{1}{c}{Bleu$_1$} & \multicolumn{1}{c}{Bleu$_2$} & \multicolumn{1}{c}{Bleu$_3$} & \multicolumn{1}{c}{Bleu$_4$} & \multicolumn{1}{c|}{ROUGE} & \multicolumn{1}{c}{Bleu$_1$} & \multicolumn{1}{c}{Bleu$_2$} & \multicolumn{1}{c}{Bleu$_3$} & \multicolumn{1}{c}{Bleu$_4$} & \multicolumn{1}{c}{ROUGE} \\ 
\cmidrule[1pt]{2-11}
G2S  & 24.51 & 15.71 & 11.19 & 8.70 & \textbf{24.84} & \textbf{23.26} & 14.54 & 10.21 & 7.84 & 22.89\\
T2S  & 22.90 & 14.55 & 10.42 & 8.14 & 23.42 & 22.14 & 13.88 & 9.85 & 7.56 & 22.50\\
TG2S (Ours) & \textbf{24.60} & \textbf{16.20} & \textbf{11.68} & \textbf{8.97} & 24.82 & 22.97 & \textbf{14.71} & \textbf{10.59} & \textbf{8.19} & \textbf{23.45}\\
\bottomrule
\end{tabular}
\caption{Back translation results obtained from the generative models when using only manual features. Our proposed model has the highest scores in almost all metrics compared to the models using only gloss or text.}
\label{tab:handsbody}
\end{table*}

\begin{table*}[!ht]
\small
\centering
\begin{tabular}{c|ccccc|ccccc}
\toprule
 \multirow{2}{*}{\textbf{Model}} & \multicolumn{5}{c|}{\textbf{Dev Set}} &  \multicolumn{5}{c}{\textbf{Test Set}} \\
& \multicolumn{1}{c}{Bleu$_1$} & \multicolumn{1}{c}{Bleu$_2$} & \multicolumn{1}{c}{Bleu$_3$} & \multicolumn{1}{c}{Bleu$_4$} & \multicolumn{1}{c|}{ROUGE} & \multicolumn{1}{c}{Bleu$_1$} & \multicolumn{1}{c}{Bleu$_2$} & \multicolumn{1}{c}{Bleu$_3$} & \multicolumn{1}{c}{Bleu$_4$} & \multicolumn{1}{c}{ROUGE} \\ 
\cmidrule[1pt]{2-11}
G2S & 16.11 & 8.77 & 5.97 & 4.49 & 16.19 & 16.29 & 9.20 & 6.37 & 4.93 & 16.73\\
T2S & 15.65 & 8.35 & 5.76 & 4.44 & 15.65 & 14.12 & 7.76 & 5.53 & 4.39 & 14.82\\
TG2S & \textbf{17.25} & \textbf{10.17} & \textbf{7.04} & \textbf{5.32} & \textbf{17.85} & \textbf{17.18} & \textbf{10.39} & \textbf{7.39} & \textbf{5.76} & \textbf{17.64}\\
\bottomrule
\end{tabular}
\caption{Back translation results obtained from the generative models when using manual features and facial landmarks and AUs. Our proposed model has the highest scores in all metrics compared to the models using only gloss or text. }
\label{tab:auc}
\end{table*}

\section{Computational Experiments}
\subsection{Features}
We extract three different types of features from the PHOENIX14T dataset: skeleton joint coordinates, facial landmark coordinates, and facial action unit intensities. We use OpenPose~\cite{openpose} to extract skeleton poses from each frame and use for our experiments the coordinates of 50 joints which represent the upper body, arms, and hands, which we will start referring to as ``manual features''. 
We also use OpenFace~\cite{baltrusaitis2018openface} to extract 68 facial landmarks as well as 17 facial action units (AUs) shown in Figure~\ref{fig:facs} to describe ``facial features''.

\subsection{Baseline Models}
We will compare the performance of our proposed model (TG2S) with two Progressive Transformers~\cite{saunders2020progressive}, one using gloss only to produce sign poses (G2S), and one that uses text only (T2S).
We train each model only with manual features and also with the combination of manual and facial features through concatenation.

\subsection{Evaluation Methods}

In order to automatically evaluate the performance of our model and the baseline models, we use back translation suggested by~\cite{saunders2020progressive}. For that purpose we use the Sign Language Transformer (SLT)~\cite{camgoz2020sign} which translates sign poses into text and computes BLEU and ROUGE scores between the translated text and the original text. As the original SLT was designed to receive video frames as input, we modified the architecture to enable the processing of skeleton poses and facial features. 

\begin{figure*}
    \centering
    \includegraphics[width=12cm]{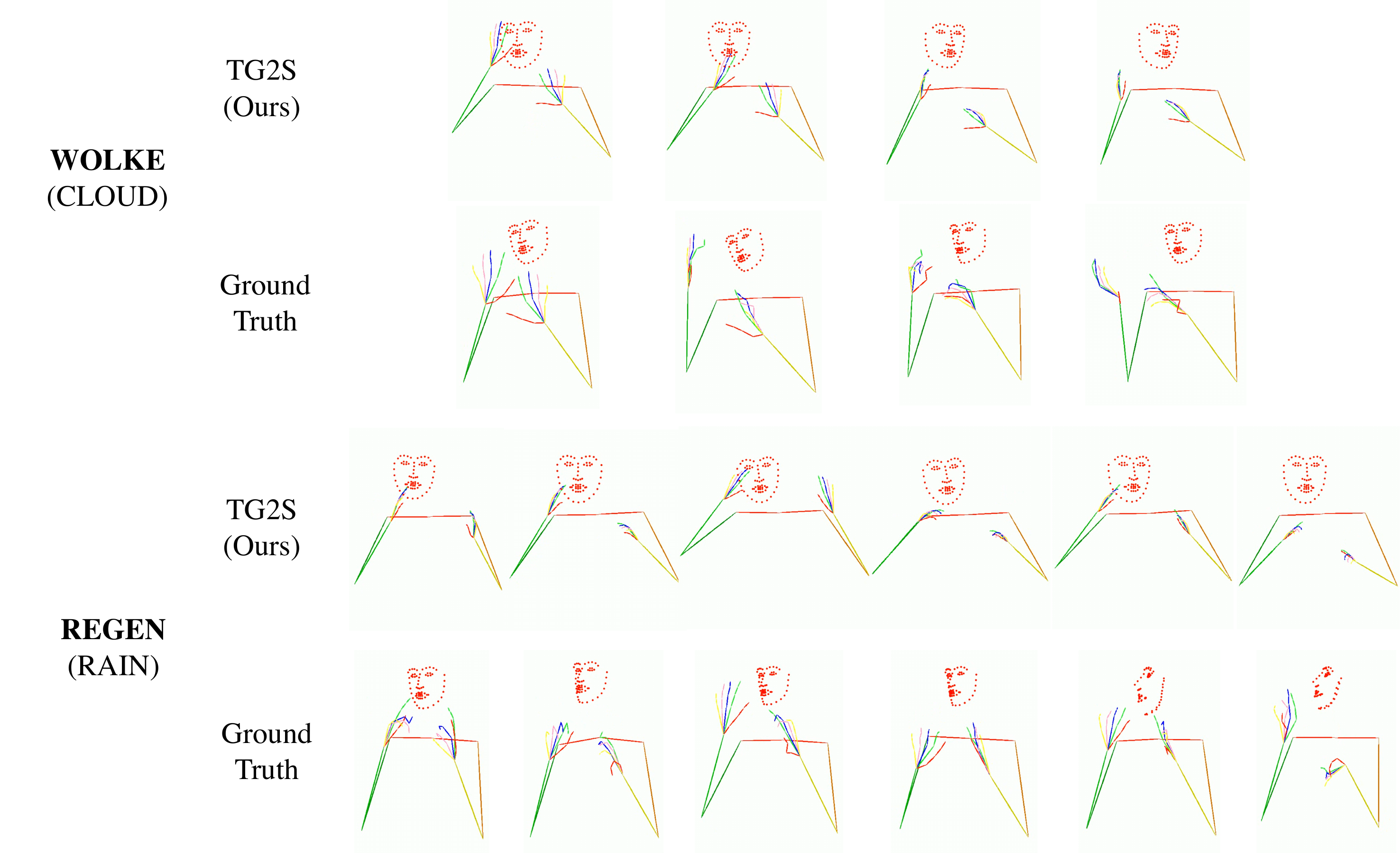}
    \caption{Comparison of the ground truth and the generated poses with our proposed dual encoder model for the gloss annotations \textsc{cloud} and \textsc{rain}. The upper example shows that the predictions captured the correct hand shape, orientation, and movement of the sign \textsc{cloud}. In the lower example it is visible that the predictions captured the repeating hand movement meaning \textsc{rain}. Although at first glance the hand orientation seems not correct, it is a slight variation which still is correct.}
    \label{fig:good_examples}
\end{figure*}

\begin{figure}
    \centering
    \includegraphics[width=0.9\linewidth]{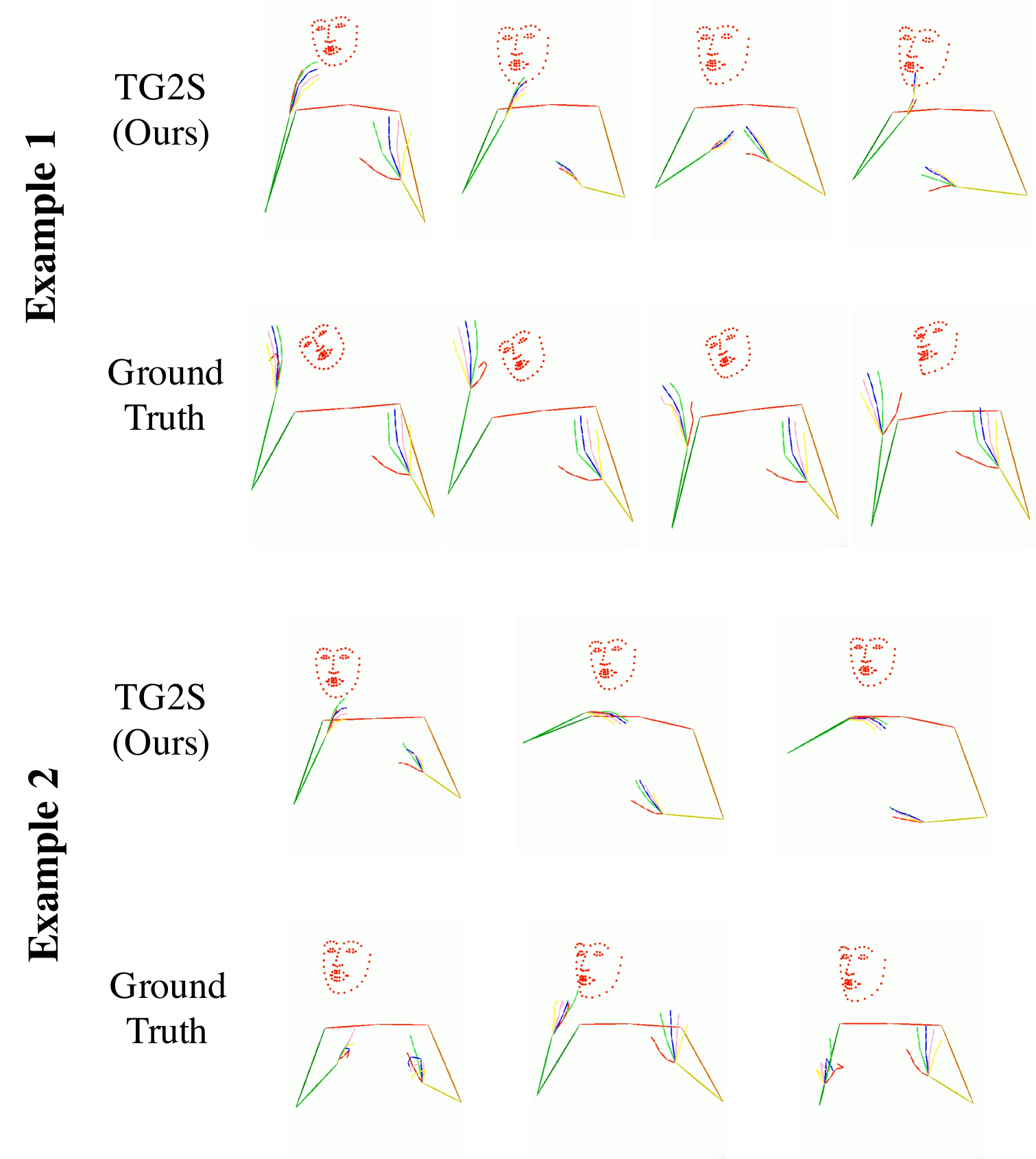}
    \caption{Examples in which our model failed to generate the correct phonology of signs. Example 1 depicts inaccuracies in hand shape, orientation, and movement. Example 2 shows the difficulty of the model to capture pointing hand shapes.}
    \label{fig:bad_examples}
\end{figure}

\section{Results}
\subsection{Quantitative Results}

Table~\ref{tab:slt} shows how well the SLT model performs the translation from ground truth sign poses to text when trained and evaluated with the PHOENIX14T dataset. The results show the highest BLEU scores are achieved when training the SLT model only with skeleton joints from hands and upper body, presenting a BLEU-4 score of 11.32 for test set. 
When facial AUs are added to the hands, body, and face features, the difference from using manual data only is slightly lower, being BLEU-4 of 10.61. 

In Table~\ref{tab:handsbody} the results of using hands and body joint skeleton as sole input to the baseline models and our proposed model are shown. We can see that our proposed model TG2S shows the highest BLEU-4 scores of 8.19 in test set, compared to 7.84 for G2S and 7.56 for T2S.

Table~\ref{tab:auc} presents the results of including facial landmarks as well as facial AUs with body and hands skeleton joints as input. Also here we can see that our proposed model outperforms the baseline models showing BLEU-4 score of 5.76 in test set. G2S obtained BLUE-4 score of 6.37 and T2S 5.53.

We see in Tables~\ref{tab:handsbody} and ~\ref{tab:auc} that G2S obtained higher scores than T2S. Given that gloss annotations fail to encode the richness of meaning in signs, it appears the smaller vocabulary helps the model achieve higher scores by neglecting information otherwise described in text. Our proposed model is able to obtain better results than G2S by making a compromise of using information from gloss, text, and their similarities and differences. 
We also can see in both tables that the inclusion of facial information reduces the overall scores. We believe that this might be the case due to the diverse range of facial expressions possible. We cannot directly compare the results of Table~\ref{tab:handsbody} and ~\ref{tab:auc} as two different SLT models were used to compute the BLEU scores.

\subsection{Qualitative Results}

Figure~\ref{fig:good_examples} shows the visual quality of our models prediction when using manual and facial information. Both examples show that the predictions captured the hand shape, orientation and movement from ground truth. In the bottom example for \textsc{rain}, the predictions were even able to capture the repetitive hand movement symbolizing falling rain. What can also be noted is that the ground truth is not perfect. In both examples unnatural finger and head postures can be seen. In addition, ground truth is not displaying movements of the eyebrows and mouth in the expected intensities.

Figure~\ref{fig:bad_examples} shows situations in which the predictions failed to represent the correct phonology of signs. In the first example we see that hand shape, orientation, and position are not correct. The predictions of our models also fail to capture pointing hand shapes as shown in example 2.
\section{Discussion and Conclusion}

In this work, for the first time, we attempt to augment contextual embeddings for sign language by learning a joint meaning representation that includes fine-grained facial expressions. Our results show that the proposed semantic representation is richer and linguistically grounded.

Although our proposed model helped bridge the loss of information by taking into account text, gloss, and their similarities and differences, there are still several challenges to be tackled by a multidisciplinary scientific community.

Complex hand shapes with pointing fingers are very challenging to generate. The first step to improve the generation of the fingers is in improving methods to recognize finger movements more accurately. Similarly, we need tools that are more robust in detecting facial expressions even in situations of occlusion. 
We also realize that SLG models are overfitting specific sign languages instead of learning a generalized representations of signs.

We chose to work with a German sign language since that is the only dataset with gloss annotation that could help us study our hypotheses. The How2Sign dataset \cite{duarte2021how2sign} is a feasible dataset for ASL, but it does not allow any model to extract facial landmarks, facial action units or  facial expression from the original video frames since the faces are blurred.
In the future, we hope to see new datasets with better and more diverse annotations for different sign languages that would allow the design of natural and usable sign language generation system.

\section*{Acknowledgements}
We want to thank Stella AI LLC for supporting Carla Viegas with computational resources and funding.
This project was partly supported by the University of Pittsburgh Momentum fund for research towards reducing language obstacles that Deaf students face when developing scientific competencies. We also acknowledge the Center for Research Computing at the University of Pittsburgh for providing part of the required computational resources.
The author affiliated with Gallaudet University was partly supported by NSF Award IIS-2118742.





\bibliography{anthology,custom}




\end{document}